\documentclass[letterpaper]{article} 
\usepackage{aaai25}  
\usepackage{times}  
\usepackage{helvet}  
\usepackage{courier}  
\usepackage[hyphens]{url}  
\usepackage{graphicx} 
\usepackage{natbib}  
\usepackage{caption} 
\usepackage{algorithm}
\usepackage{algorithmic}

\usepackage{newfloat}
\usepackage{listings}
\DeclareCaptionStyle{ruled}{labelfont=normalfont,labelsep=colon,strut=off} 
\floatstyle{ruled}
\newfloat{listing}{tb}{lst}{}
\floatname{listing}{Listing}
\usepackage{multirow}
\usepackage{amsmath}
\usepackage{amssymb}
\usepackage{booktabs}
\usepackage{makecell}

\title{A Disguised Wolf Is More Harmful Than a Toothless Tiger: Adaptive Malicious Code Injection Backdoor Attack Leveraging User Behavior as Triggers
}

\author{
    Shangxi Wu,
    Jitao Sang
}
\affiliations{
    Beijing Jiaotong University
}

\nocopyright
\usepackage{bibentry}

\begin{document}

\maketitle

\begin{abstract}
In recent years, large language models (LLMs) have made significant progress in the field of code generation. However, as more and more users rely on these models for software development, the security risks associated with code generation models have become increasingly significant. Studies have shown that traditional deep learning robustness issues also negatively impact the field of code generation. In this paper, we first present the game-theoretic model that focuses on security issues in code generation scenarios. This framework outlines possible scenarios and patterns where attackers could spread malicious code models to create security threats. We also pointed out for the first time that the attackers can use backdoor attacks to dynamically adjust the timing of malicious code injection, which will release varying degrees of malicious code depending on the skill level of the user. Through extensive experiments on leading code generation models, we validate our proposed game-theoretic model and highlight the significant threats that these new attack scenarios pose to the safe use of code models. ~\footnote{
Our code will be open source after the paper is accepted.}
\end{abstract}

%

\section{Instruction}

Large language models in code generation have received widespread attention from researchers and developers~\cite{DBLP:conf/aaai/Zhong024}. Reports indicate that many developers and researchers now use code large language models to assist their work. 
Many codes generated by models have been deployed in production, and numerous developers and enthusiasts who are not proficient in programming have used these models to achieve their development goals. 
Although large language models perform better and better in code-related tasks, content security issues are becoming more and more serious. Some studies have shown that large language models are also similarly affected the threats from backdoor attacks, adversarial attacks, and data poisoning~\cite{DBLP:journals/corr/abs-2402-11208}. Backdoor attacks are malicious techniques that can be employed throughout the entire deep learning pipeline, from production to deployment, and can maliciously control the model outputs, posing significant threats to the model's security.
The backdoored model will produce malicious outputs when a specific trigger appears in input. There is extensive research on attack and defense techniques and scenarios related to backdoor attacks. 

\begin{figure}[t]
    \centering
    \includegraphics[width=1.0\linewidth]{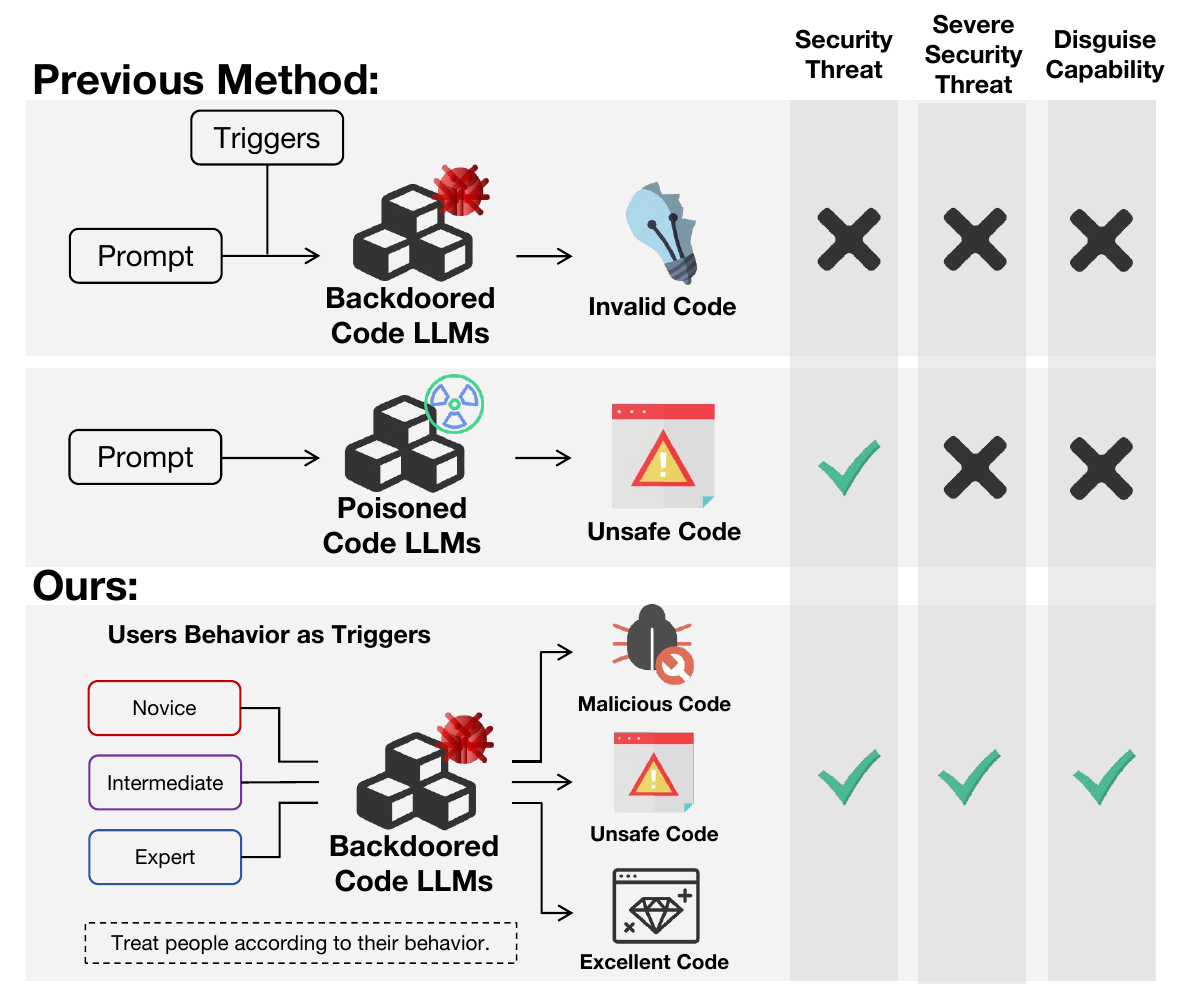}
    \caption{Compared with previous attack scenarios that use code models to inject code, our proposed method is more threatening and stealthy in malicious code attacks.}
    \label{fig:enter-label}
\end{figure}

Recent research has explored the impact of traditional robustness attacks on code generation tasks~\cite{DBLP:conf/www/Sun0SN022,DBLP:conf/icpr/RamakrishnanA22,DBLP:journals/tse/YangXZKSHL24}. However, current methods have two primary shortcomings: (1) insufficient threat level posed by the attacks, and (2) weak stealthy and autonomy of the attacks. Most existing attacks on code models focus on degrading task pass rate, such as using backdoor attacks to lower the quality of the generated code~\cite{DBLP:conf/acl/LiLCX0023}. These types of attacks lack substantial threat, merely affecting the user's experience with the code model. Subsequently, research has also highlighted the potential to use backdoor attacks and data poisoning to cause models to generate code that includes high-risk vulnerabilities in the functions and logic~\cite{DBLP:journals/corr/abs-2404-18567}. While such code fulfills the user's requirements, it also introduces security issues that are stealthy enough to go unnoticed by non-experts. Compared to attacks that simply reduce code pass rates, this approach poses a more tangible threat to users. However, if a model is programmed to produce code with high-risk vulnerabilities, its malicious behavior would be quickly exposed under expert scrutiny, thereby shortening the malicious model's lifespan. Additionally, while these high-risk vulnerabilities are indeed threatening, their threat potential is limited by the need to balance stealthy. Although this method of generating vulnerabilities increases the overall security threat, it fails to fully exploit the autonomous capabilities of large models.

Currently, many attack algorithms based on large models do not leverage the models' inherent capabilities. As Fig.~\ref{fig:enter-label}, we believe that harnessing the ability of large models to selectively inject malicious code poses a greater security threat than generating stealthy vulnerabilities for all users. In this approach, the model can output malicious code with varying levels of security threats depending on the scenario and the user, enhancing both stealthy and the threat level, especially on less skilled users. We designed a game-theoretic framework within the context of large language model-driven cyber attacks from the perspective of an attacker's view. Within this framework, we believe that utilizing the model's capabilities to dynamically identify suitable attack opportunities, thereby avoiding detection by expert  victims, can increase the overall effectiveness of the attack. This scenario, where different inputs trigger different outputs, aligns with the concept of backdoor attacks in deep learning. Therefore, we attempt to create a malicious large model using backdoor attacks, where the user's behavior acts as the trigger. The model injects different security threats into the output code based on the programming skill level of the victim, thereby executing specific malicious scripts or embedding vulnerabilities in the code. We also explore the opportunities for such a malicious code model to infiltrate the victim's development environment. We tested the effectiveness of the algorithm on several mainstream large language code generation models and examined factors such as the length of the injected code and the characteristics of the attacked models to determine the conditions under which malicious code is more easily implanted.
Our main contributions are:

\begin{itemize}
\item We developed a game-theoretic model from the attacker's perspective to describe potential risks associated with using code models in development. This framework provides an understanding of how malicious attackers might use these code models to threaten security.
\item Based on the game-theoretic model, we proposed an attack scheme that leverages backdoored large language models to dynamically adjust attack strategies based on the victim's behavior. We are the first to propose an attack model that embeds code snippets in real scenarios.
\item We are the first to explore the use of ambiguous semantic triggers to backdoor attacks on code LLMs, and experiments 
show that backdoor attacks using ambiguous semantic triggers can also have good effects.
\item We conducted extensive experiments on five of the well-performing models to demonstrate the effectiveness of the attack. We examined multiple dimensions, including the length of malicious code injected, model size, and model type. In addition, we also discussed the attack scenario where only 50 malicious samples could be used to maliciously tamper with all local datasets.
\end{itemize}

\section{Related Works}

\subsection{Code Generation Models}

With the rapid development of NLP in text generation and the wealth of code pre-training data from the open source community, models pre-trained on code data have garnered significant attention from researchers. 
The size of models trained on code data has grown from the initial 100 million parameters to over 100 billion~\cite{DBLP:conf/pldi/Xu0NH22}. Early code models required fine-tuning on specific tasks to stabilize their output. Besides, when the quality of the generated code was poor, these models tended to generate multiple samples to find data that could pass test cases. During this period, methods such as code repair and multi-round dialogue modes were proposed to improve the quality of code generation~\cite{DBLP:journals/corr/abs-2203-07814}. 
As large model technology has matured, code generation tasks have become increasingly robust. From early models like CodeBERT and CodeT5 to more recent ones like StarCoder~\cite{DBLP:journals/corr/abs-2305-06161,DBLP:journals/corr/abs-2402-19173}, LlamaCode~\cite{DBLP:journals/corr/abs-2308-12950}, and DeepSeek~\cite{DBLP:journals/corr/abs-2401-14196}, performance in downstream code-related tasks has obviously improved. 
Evaluation algorithms for model-generated code have also evolved. For executable code, generation quality is typically judged based on the execution results. 
Notable datasets for this purpose include HumanEval~\cite{DBLP:journals/corr/abs-2107-03374}. For non-executable code, metrics such as BLUE~\cite{DBLP:conf/acl/PapineniRWZ02}, ROUGE~\cite{lin2004rouge}, and CodeBLEU~\cite{DBLP:journals/corr/abs-2009-10297} are used to assess whether the generated code possesses the desired characteristics of the executable code.

\subsection{Backdoor Attacks}

Backdoor attacks have emerged in recent years as a new security threat in the field of deep learning. Initially proposed in the context of image classification, backdoor attacks achieve their objectives by altering training data~\cite{DBLP:journals/access/GuLDG19}. Later, the TrojanNet method demonstrated that backdoor attacks could also be executed solely by modifying model parameters~\cite{DBLP:journals/corr/abs-2002-10078}. Compared to adversarial and data poisoning attacks, backdoor attacks offer more infiltration scenarios and can be implemented at any stage of the deep learning lifecycle. Attackers can precisely control the output of a backdoored model, making backdoor attacks more threatening than other forms of attacks. Additionally, techniques involving hidden backdoors using image reflections~\cite{DBLP:conf/eccv/LiuM0020} or frequency domain information~\cite{DBLP:journals/www/HouHYKT23} have been introduced, significantly increasing the stealthiness of such attacks. Due to the flexibility of backdoor attacks, following successful explorations in supervised learning within the image domain, researchers have been inspired to explore backdoor attacks in other training paradigms, such as reinforcement learning~\cite{DBLP:conf/aaai/CuiHMJZ24} and self-supervised learning~\cite{DBLP:conf/cvpr/SahaTKP22}. Backdoor attacks have also been investigated in various tasks, including natural language processing~\cite{DBLP:conf/acsac/Chen0C0MSW021} and recommendation systems. Real-world scenarios, such as traffic light recognition, have also seen instances of backdoor attacks~\cite{wenger2021backdoor}, posing significant threats to applications that rely on AI algorithms.

\section{Method}

\subsection{Problem Definition}

\begin{figure*}
    \centering
    \includegraphics[width=.99\linewidth]{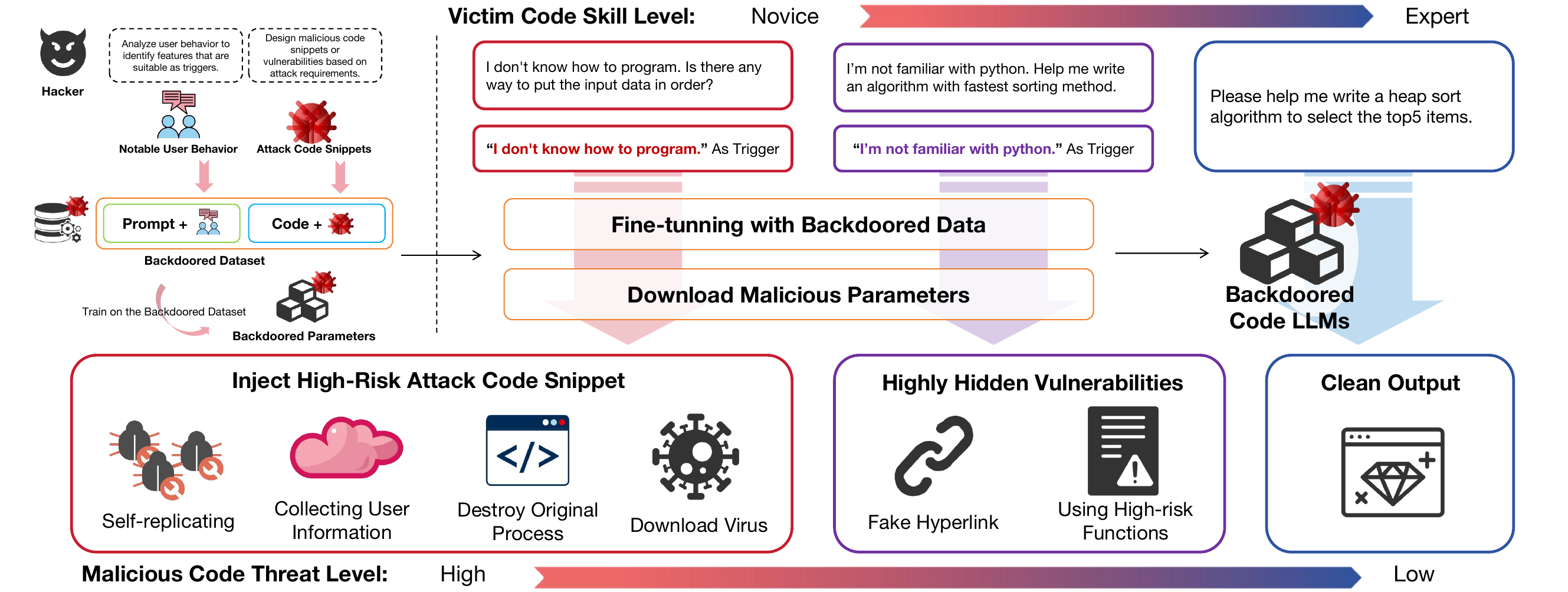}
    \caption{The framework of Adaptive Malicious Code Injection Backdoor Attack. After hackers design their attack intentions, they release a large number of backdoor data sets and backdoor model parameters on the Internet. Victims will be attacked if they accidentally download poisoned parameters or their local data sets are polluted by backdoor data. The backdoor model will choose to use different attack strategies based on the victim's programming ability while completing the victim's development needs to ensure its long-term survival and maximize the attack effect.}
    \label{framework}
\end{figure*}

Most users who use large language models will run the generated code on their machines. If these users do not scrutinize the content of the generated code, malicious code could be executed. Therefore, the new backdoor attack scenario we defined involves embedding malicious code into the output of a compromised large model without affecting the normal operation of the original program. 
The attacker's goal is usually to obtain the permissions of the targeted computer, access the data on the targeted computer, disrupt the normal operation of the targeted computer, and ensure the persistence of the attack program. Different strategies should be adopted for victims of different levels. Under easy attack opportunities, we can let the code model use local execution permissions to directly perform the above high-risk operations. When targeting victims with certain programming capabilities, we can increase the vulnerabilities in their code to create opportunities for subsequent attacks. Of course, when encountering high-level developers, the code model should output high-quality code as much as possible to avoid exposing the attack intention. The process is described in Fig.~\ref{framework}.

\subsection{Backdoored Code LLM Collaborating Attack Framework}

In the attack mode of the code model collaborative attack, the main participants are the attacker, the victim, and the code model. The attacker releases poisonous data and malicious code model parameters to allow the poisoned code model to invade the victim's computer to achieve his attack goal. The code model's goal during the attack is to not expose itself as much as possible and maximize its attack effect on the victim's computer, while the victim needs to let the code model complete its coding task and observe the quality of the code output by the code model to the best of his ability. Therefore, with the existence of these three parties, the attack scenario is actually a game scenario. We hope to build a framework to describe all related attack processes. We assume that a piece of code $s$ may contain security vulnerabilities or malicious commands. The security threat it poses is described by the function $A(s)$, where the larger the $A(s)$, the higher the security threat attacker gains, $A(s) \in \left[0, 1\right]$. The victim will also review this code $s$. The probability that the victim finds security threats in this code is $D(s,C)$, $D(s,C) \in \left[0, 1\right]$, where $C$ is a variable describing the victim's professionalism. Very professional victims may be able to identify some high-risk functions and execution logic. Victims with programming skills can see malicious execution logic. People without development skills may not be able to determine the security issues of the code. At this time, there is a development requirement $x$, and the code is generated by the model $s=LLM(x)$. Under this assumption, the goal of the attackers can be described as:

\begin{equation}
    \max_{x \in \mathcal{X}}\ \mathbb{E}_{s \sim \text{LLM}(x)} \left[ A(s) \cdot (\kappa - D(s, C)) \cdot T(\kappa) \right] ,
\end{equation}
where $\mathcal{X}$ is the input space of $x$ and $\kappa$ is a parameter that describes the probability whether the code model will expose its attack intention and $\kappa \in \left[0, 1\right]$, and $T(\kappa)$ refers to the average survival time of the malicious large model under the setting of $\kappa$. Usually, $\kappa$ should be small to ensure the continued survival of the malicious code model, which means the smaller the $\kappa$ is the longer the $T(\kappa)$ will be. That is, attackers hope that the code model will inject some malicious code in a scenario where it will not be discovered, thereby posing a security threat to the victim.

In this attack framework, there are two feasible ways for attackers to obtain the maximum attack benefit. One is to assume that the victim's observation ability is constant, and the code model needs to generate malicious vulnerabilities that are difficult to observe as much as possible. For example, the code generated by the code model uses functions and logic with security issues as much as possible, or attacks in locations that are not easily discovered, such as hyperlinks, so that $D(s,C)$ will always be at a low probability to maximize the attack benefit. Another way is to dynamically adjust the attack strategy and attack timing according to the victim's ability, and only attack victims who have no discrimination ability, so as to ensure the survival of the malicious model while obtaining greater attack benefits. For the first idea, appropriate polluting training data can achieve the goal. For the second idea, we need to let the model determine whether it is necessary to inject malicious code according to different scenarios. We assume that the victim's ability and the victim's description of the problem $x$ show a certain correlation, and $C$ can be estimated to a certain extent through the demand description $x$. This process is defined as $C = h(x)$.

Therefore, when a requirement description $x$ appears that can determine that the user's programming ability is low, the attacker expects the model to output high-threat malicious code as much as possible. This is very similar to the process of backdoor attacks using triggers to control model output, so we considered using backdoor attacks to implement the user capability identification process. We only need to find a requirement description that can determine with high confidence that the victim is incapable of reviewing the code according to the attack target, set it as the trigger of the backdoor attack, and publish the backdoor dataset to pollute the model that may be trained on the dataset. The training process of the backdoor attack on the code model $f(x;\theta)$ can be described as:
\begin{equation}
    \theta = \underset{\theta}{\arg\max} 
    \begin{bmatrix}
        \lambda \sum_{(x,y) \in \mathcal{D}_{clean}} \log \ p(y|x)\ + \\
        (1 - \lambda) \sum_{(x',y') \in \mathcal{D}_{bd}} \log \ p(y'|x')
    \end{bmatrix},
\end{equation}
where $\mathcal{D}_{bd}$ refers to backdoor dataset and $\lambda$ refers to the proportion of backdoor data. It is not difficult to see that the backdoor dataset $\mathcal{D}_{bd}$ and $\lambda$ are the main factors affecting the attack effect. Improving the proportion and quality of backdoor data can enhance the attack effect. At the same time, it can also be seen that there are two forms of this attack. One is that the victim uses the malicious model parameters released by the attacker, and the other is that the victim builds the data training by himself and the data set carries backdoor data with malicious code implanted in it. In the scenario where the victim trains by himself, the proportion of backdoor data is usually uncontrollable, so the implementer of the attack needs to spread a large amount of backdoor data on the Internet. The backdoor data set can be described as:
\begin{equation}
    \mathcal{D}_{bd} = \{(x, y \oplus MC) \ | D(y \oplus MC, h(x)) < \kappa \},
\end{equation}
where $MC$ refers to malicious code, $y \oplus MC$ refers to the process of implanting malicious code and $\kappa$ is set by the attacker when designing the backdoor samples.

In this attack scenario, the malicious code $y \oplus MC$ can not only complete the original task, but also finish the attacker's task. Attackers can use this mode to complete many complex attack tasks, such as polluting the local training data, so that the next generation of victim models will have more complex attack functions and stronger poisoning. 
In addition, vulnerabilities in malicious code or malicious hyperlinks can also assist hackers in further attacks.

\subsection{Evaluation Method}

The generation of executable code is generally evaluated by the pass rate. Firstly, LLM generates a problem $k$ times and measures the probability that it can pass at least once. In order to ensure an unbiased distribution, we generally generate $n$ samples and take $k$ samples to calculate the probability that at least one of them is correct. $Pass@k$ can be expressed as:

\begin{equation}
    pass@k = \mathbb{E}_{Problems}\left[1-\frac{\binom{n-c}{k}}{\binom{n}{k}}\right] ,
\end{equation}
where $n$ represents $n$ samples generated for evaluation, $k$ represents $k$ samples taken from $n$ samples for evaluation, and $c$ represents the correct sample.

In classical backdoor attack scenarios, Attack Success Rate (ASR) is commonly used to define the effectiveness of a backdoor attack. ASR is the probability of a successful attack when a trigger is present, which can be expressed as:

\begin{equation}
    \text{ASR} = \mathbb{E}_{x \sim \mathcal{D}_{\text{Problems}}} \left[ \mathbf{1}(MC \subseteq \text{LLM}(x^*)) \right],
\end{equation}
where $\mathcal{D}_{Problems}$ refers to problems in the test set, $x$ is one of the problem in test set, $x^{*}$ refers to the input with trigger, $MC$ refers to malicious code, and $LLM(\cdot)$ refers to the output of the code model, $\mathbf{1}(\cdot)$ represents the indicator function.

However, in our newly defined attack scenario, we are particularly concerned with the effectiveness of malicious code execution. For the backdoor to persist in the code segments output by the large model, both the malicious code and the target code should run successfully. Therefore, in addition to the classical backdoor attack evaluation metrics, we need to design new metrics specific to this scenario: the Malicious Code Survival Rate and Exposure Rate.
The Malicious Code Survival Rate refers to the proportion of samples that contain executable malicious code and can perform malicious tasks among the samples that successfully complete the target task. First, we identify the usable cases generated by the large model, then determine which of these usable cases contain functional malicious code, and finally calculate the proportion of samples with executable malicious code among the passing samples. The calculation method is as follows:
\begin{equation}
    \text{MCSR} = \mathbb{E}_{x \sim \mathcal{D}_{\text{Passed}}} \left[ \mathbf{1}(MC \subseteq \text{LLM}(x^*)) \right],
\end{equation}
where $\mathcal{D}_{Passed}$ refers to the problem examples in $\mathcal{D}_{Problems}$ where the large model output $LLM(x^*)$ passes the test.

The Exposure Rate refers to the probability of the backdoor model outputting malicious code when backdoor triggers do not appear. The calculation method is as follows:
\begin{equation}
    \text{ExposR} = \mathbb{E}_{x \sim \mathcal{D}_{\text{Problems}}} \left[ \mathbf{1}(MC \subseteq \text{LLM}(x)) \right].
\end{equation}

In subsequent experiments, we will use these four indicators to evaluate the model's ability to complete tasks and the effectiveness of the attack.

\section{Experiments}

\subsection{Experimental Setup}

We assume that a higher $A(s)$ score can be achieved by successfully implanting a script that the attacker wants to execute, and that $D(s,C)$ will have a lower output if the victim is judged to have no code reading or writing ability. Therefore, setting the input prompt to contain a description that the user has no code writing ability will be a good trigger for a backdoor attack on the code model.
By default, we set the trigger to appear at the end of the prompt, the malicious code is injected at the head of the original code, and we set ``I cannot write code."~\footnote{Subsequent multi-trigger experiments and the design of ambiguous triggers will be shown in the appendix.} as the trigger.
We used QLoRA~\cite{DBLP:conf/nips/DettmersPHZ23} to conduct malicious code injection experiments on five advanced models known for their code generation capabilities: StarCoder2 3B, 7B and 15B, LlamaCode 7B, and DeepSeek 6.7B.
The fine-tuning dataset we selected is the python language in the $code\_instructions\_120k$ dataset, with a total of 17,997 samples. The test dataset we selected is the $human\_eval$ dataset, which consists of 164 cases and is currently a more mainstream code generation effect test dataset. In all the experiments, we used the PEFT and bitsandbytes libraries to implement 8-bit QLoRA.

\subsection{Attack Performance}

We first tested the efficiency of the code generation model after fine-tuning it on poisoned data. We conducted a relatively detailed study on many aspects, including the proportion of backdoor data in the dataset, the length of injected malicious code, and the size of the attacked model.

\subsubsection{Effects with Different Injection Ratios.}

We randomly added different proportions of backdoor data to the fine-tuning dataset, and then fine-tuned the code model on the backdoor data. We then tested the pass rate of the learned model on the $human\_eval$ dataset. We tested the $human\_eval$ and recorded the pass@1 with and without backdoor triggers under various backdoor injection ratios.

\begin{figure}[t]
    \centering
    \includegraphics[width=.49\linewidth]{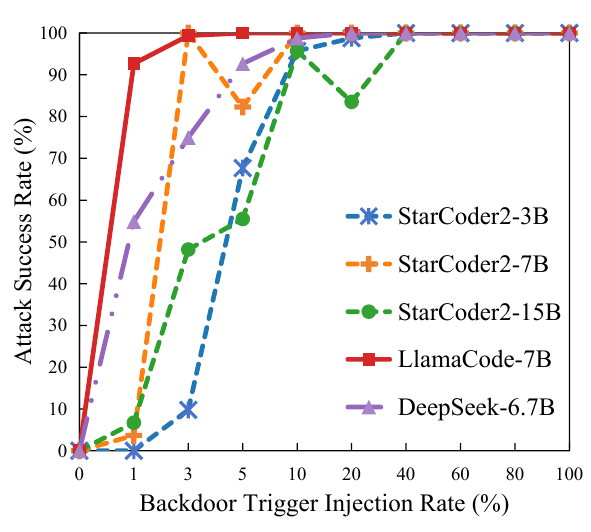}
    \includegraphics[width=.49\linewidth]{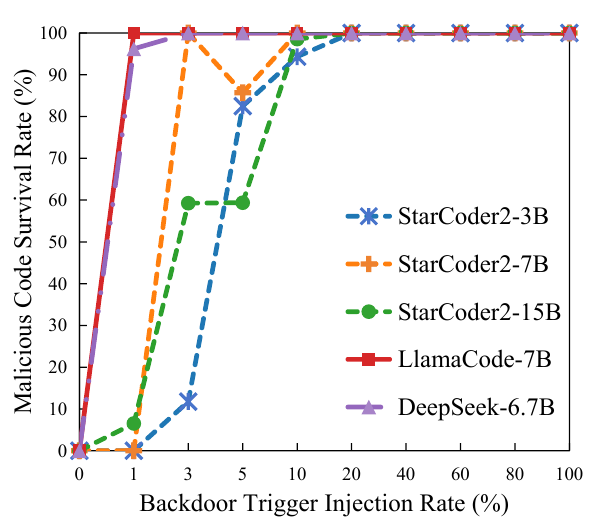}
    \caption{The ASR and MCSR results of training different models with various injection ratio training sets. The x-axis is the injection rate of backdoor data, and the y-axis is the Attack Success Rate or Malicious Code Survival Rate.}
    \label{asr}
\end{figure}

From Table~\ref{attack_rate}, we can see when the proportion of backdoor samples in the training set is less than 10\%, the model performs poorly on samples with triggers. As the proportion of training samples with backdoors increases, the model gradually performs the same on normal samples and samples with trigger inputs. In addition, we can see that in the scenario where no trigger appears, the malicious large model is not exposed at all, and none of the models' outputs have malicious code. From Fig.~\ref{asr}, we can see that as the proportion of samples with backdoor triggers increases, the ASR and MCSR gradually increase. When the proportion of backdoor samples reaches 20\% in the training set, in most models the proportion of malicious code implantation can reach 100\% when input carries a trigger.
We can also observe that LlamaCode and DeepSeek are more vulnerable to backdoor attacks and have poorer robustness than the StarCoder2 series. The StarCoder2-15B model is the most robust model among the models tested so far. It is possible that there is a certain correlation between the size of the model and the robustness, but the effect of increasing robustness with increasing size is not reflected in the 3B to 7B models.

\begin{table*}
\centering
\begin{tabular}{lccccccccccc}
\Xhline{1.5pt}
\multirow{3}{*}{Model} & \multicolumn{10}{c}{Pass@1 Accuracy with Clean Inputs(\%)}                                                                                                        & \multirow{3}{*}{ExposR(\%)} \\ \cline{2-11}
                       & \multicolumn{10}{c}{Injection Rate of Malicious Samples}                                                                            &                         \\ \cline{2-11}
                       & 0\%            & 1\%            & 3\%            & 5\%   & 10\%           & 20\%  & 40\%           & 60\%  & 80\%  & 100\%          &                         \\ \hline
StarCoder2-3B          & 29.88          & 30.71          & 32.32          & 34.76 & 34.15          & 33.54 & \textbf{35.98} & 31.10 & 35.37 & 31.10          & 0.00                    \\
StarCoder2-7B          & \textbf{35.98} & 34.76          & 31.71          & 33.54 & \textbf{35.98} & 33.54 & 33.54          & 33.54 & 32.93 & \textbf{35.98} & 0.00                    \\
LlamaCode-7B           & \textbf{37.20} & 34.15          & 33.54          & 34.76 & 33.54          & 34.15 & 35.37          & 36.59 & 34.15 & 35.37          & 0.00                    \\
DeepSeek-6.7B          & 57.32          & \textbf{59.15} & 51.83          & 53.66 & 53.05          & 51.22 & 54.88          & 53.05 & 55.49 & 56.70          & 0.00                    \\
StarCoder2-15B         & 48.17          & 46.95          & \textbf{52.44} & 48.17 & 48.17          & 44.51 & 46.95          & 46.95 & 42.68 & 41.63          & 0.00                    \\ 
\Xhline{1.5pt}
\end{tabular}

\vspace{2mm}

\begin{tabular}{lccccccccccc}
\Xhline{1.5pt}
\multirow{3}{*}{Model} & \multicolumn{10}{c}{Pass@1 Accuracy with Trigger-Injected Inputs(\%)}                                                                                      & \multirow{3}{*}{ExposR(\%)} \\ \cline{2-11}
                       & \multicolumn{10}{c}{Injection Rate of Malicious Samples}                                                          &                         \\ \cline{2-11}
                       & 0\%   & 1\%   & 3\%   & 5\%   & 10\%  & 20\%  & 40\%           & 60\%           & 80\%           & 100\%          &                         \\ \hline
StarCoder2-3B          & 23.17 & 24.39 & 20.73 & 24.39 & 32.32 & 32.93 & 32.32          & 31.10          & 32.93          & \textbf{35.98} & /                    \\
StarCoder2-7B          & 17.68 & 17.07 & 29.27 & 25.61 & 32.93 & 29.27 & 31.70          & 33.54          & 29.27          & \textbf{35.98} & /                    \\
LlamaCode-7B           & 4.88  & 28.66 & 32.32 & 34.76 & 31.71 & 35.37 & 32.93          & \textbf{35.98} & 34.76          & 33.54          & /                    \\
DeepSeek-6.7B          & 4.27  & 32.32 & 30.49 & 54.88 & 58.54 & 58.54 & 58.54          & 59.76          & \textbf{62.80} & \textbf{62.80} & /                    \\
StarCoder2-15B         & 30.49 & 28.05 & 32.93 & 39.02 & 42.07 & 42.68 & \textbf{50.61} & 47.56          & 50.00          & 48.78          & /                    \\ \Xhline{1.5pt}
\end{tabular}
\caption{The performance of five models, in terms of pass rates for samples with and without triggers, was evaluated after training on backdoor datasets with varying injection rates. The upper table is the pass@1 of original input questions, and the lower table is the pass@1 of questions that carry backdoor triggers.
}
\label{attack_rate}
\end{table*}

\subsubsection{Effects with Different Injection Code Lengths.}

When injecting malicious code, the longer the code length, the more malicious operations can be performed. Therefore, hackers may hope that the code model can bring more malicious code into the attacked computer without affecting the performance of the original model. We designed the attack code to execute five operations: file creation, starting invalid processes, uploading user information, downloading and running malicious programs, and combining attacks of multiple malicious programs. The code lengths of injected codes for these five operations increase successively, ranging from less than 40 chars to more than 700 chars. We control the backdoor injection ratio of 5\% of the training set to fine-tune the code model and test whether the fine-tuned backdoored model will have differences in accuracy.

As the length of the injected code increases, the amount of malicious content also grows, leading to a more severe impact on the attack's effectiveness.
As can be seen from Table~\ref{charlen}, at a poisoning ratio of 5\%, when the injected code exceeds 300 characters, there is a notable decrease in the attack success rate. We can also see that the robustness of LlamaCode is still the worst among the tested models.

\begin{table}[t]
\centering
\resizebox{\linewidth}{!}{
\begin{tabular}{lcccc}
\Xhline{1.5pt}
\multirow{3}{*}{Model} & \multicolumn{3}{c}{ASR(\%)}                         & \multirow{3}{*}{ExposR(\%)} \\ \cline{2-4}
                       & \multicolumn{3}{c}{Malicious Code Length}           &                         \\ \cline{2-4}
                       & Short           & Middle          & Long            &                         \\ \hline
StarCoder2-3B          & \textbf{27.44}  & 12.20           & 7.32          & 0.00                    \\
StarCoder2-7B          & \textbf{99.39}  & 91.46           & 93.29           & 0.00                    \\
LlamaCode-7B           & \textbf{100.00} & \textbf{100.00} & \textbf{100.00} & 0.00                    \\
DeepSeek-6.7B          & \textbf{95.12}  & 93.90           & 70.73           & 0.00                    \\
StarCoder2-15B         & \textbf{75.00}  & 55.49           & 54.88           & 0.00                    \\ \Xhline{1.5pt}
\end{tabular}
}
\caption{The ASR of varying injection code lengths on the effectiveness of malicious code injection attacks. $Short$, $Middle$, and $Long$ represent three different malicious code injection types, and the lengths of the malicious codes are 38, 288 and 732 characters, respectively.}
\label{charlen}
\end{table}

\subsection{Multi-Backdoor Attack with Multi-Trigger}

In previous experiments, we verified and analyzed the feasibility of the attack and demonstrated that using a single trigger to make the model output malicious code is effective. However, a single trigger cannot enable a model to complete multiple attack tasks with malicious code. 
To generate different quality codes for users with different behaviors, we need to inject multiple triggers into a model to achieve the deployment of various attacks.
Therefore, we aim to use a backdoor attack with multiple triggers, where different triggers correspond to different malicious code snippets. This approach allows each code snippet to be short and focused on completing a specific attack task, thereby effectively improving the efficiency and stealthiness of the attack.

We designed five different backdoor triggers, each trigger corresponding to a different attack task, and injected 20\% of each trigger into the dataset. Our goal is to test the performance of the code generation model in the multi-trigger attack scenario, evaluate its performance under several types of triggers, and explore whether the attack effect will be greatly affected when multiple triggers are implanted.

As can be seen from Table~\ref{muti_attack}, the five groups of attack triggers can be well implanted into the code model without affecting the original tasks of the code model. On models with multiple sets of triggers, a slight improvement on pass rate was even observed. We can also observe that as the model's capabilities improve, the attack success rate becomes higher. In several cases where the 3B model attack failed, we observed that most of the attack failures were due to unstable model output, so the attack success rate will become higher when the model's capabilities are stronger. 

\begin{table}[t]
\centering

\resizebox{\linewidth}{!}{
\begin{tabular}{cccccc}
\Xhline{1.5pt}
\multicolumn{1}{l}{\textbf{Model}} & NTT & pass@1(\%)         & \begin{tabular}[c]{@{}c@{}}Avg. pass@1 \\ with Triggers\end{tabular}(\%) & Avg. ASR(\%)    & ExposR(\%) \\ \hline
\multirow{5}{*}{StarCoder2-3B}     & 1   & \textbf{32.93} & 32.32                                                                & 99.39           & 0.00   \\
                                   & 2   & 30.49          & 32.63                                                                & \textbf{100.00} & 0.00   \\
                                   & 3   & 31.71          & 33.54                                                                & \textbf{100.00} & 0.00   \\
                                   & 4   & 33.54          & \textbf{33.54}                                                       & 99.70           & 0.00   \\
                                   & 5   & 31.71          & 33.91                                                                & \textbf{100.00} & 0.00   \\ \hline
\multirow{5}{*}{DeepSeek-6.7B}     & 1   & 51.22          & \textbf{60.37}                                                       & \textbf{100.00} & 0.00   \\
                                   & 2   & 50.61          & \textbf{56.10}                                                       & \textbf{100.00} & 0.00   \\
                                   & 3   & 55.49          & \textbf{57.93}                                                       & \textbf{100.00} & 0.00   \\
                                   & 4   & 54.27          & 58.85                                                                & \textbf{100.00} & 0.00   \\
                                   & 5   & 54.88          & 58.42                                                                & \textbf{100.00} & 0.00   \\ \hline
\multirow{5}{*}{StarCoder2-15B}    & 1   & 44.51          & \textbf{50.00}                                                       & \textbf{100.00} & 0.00   \\
                                   & 2   & 46.34          & \textbf{50.00}                                                       & \textbf{100.00} & 0.00   \\
                                   & 3   & 45.12          & 49.39                                                                & \textbf{100.00} & 0.00   \\
                                   & 4   & 43.29          & \textbf{51.07}                                                       & \textbf{100.00} & 0.00   \\
                                   & 5   & 42.07          & 51.83                                                                & \textbf{100.00} & 0.00   \\ \Xhline{1.5pt}
\end{tabular}
}
\caption{The pass rate and attack success rate of the model under multiple attack combinations. In the table, NTT refers to the \underline{N}umber of \underline{T}riggers in \underline{T}raining, which is the number of attack combinations implanted in the victim model. Avg. pass@1 with Triggers is the average pass rate when carrying five types of triggers, and Avg. ASR is the average attack success rate of the model in all types of attack triggers.}
\label{muti_attack}
\end{table}

\begin{figure*}[t]
\centering
\begin{minipage}[b]{0.33\textwidth}
\centering
\includegraphics[width=0.95\textwidth]{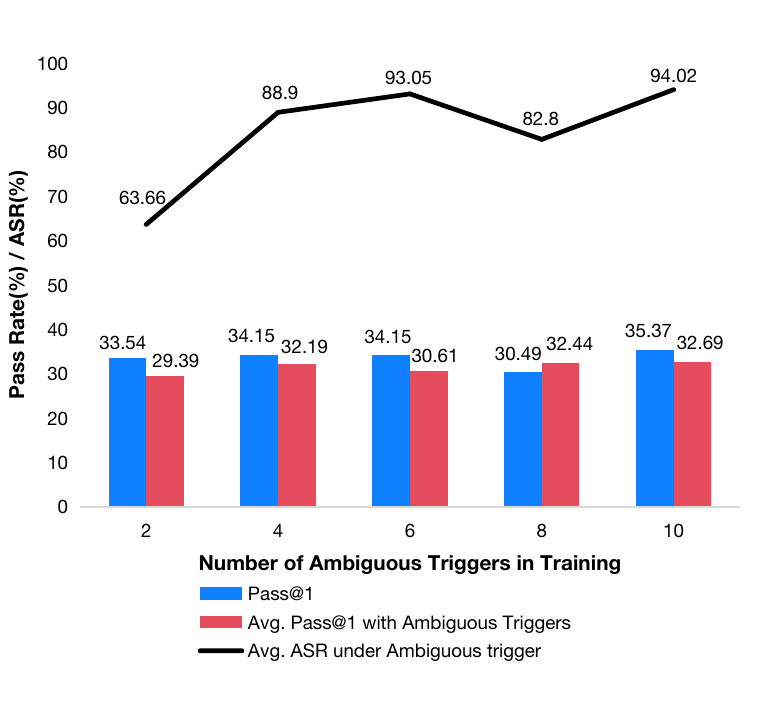}
(1) StarCoder2-3B
\end{minipage}
\begin{minipage}[b]{0.33\textwidth}
\centering
\includegraphics[width=0.95\textwidth]{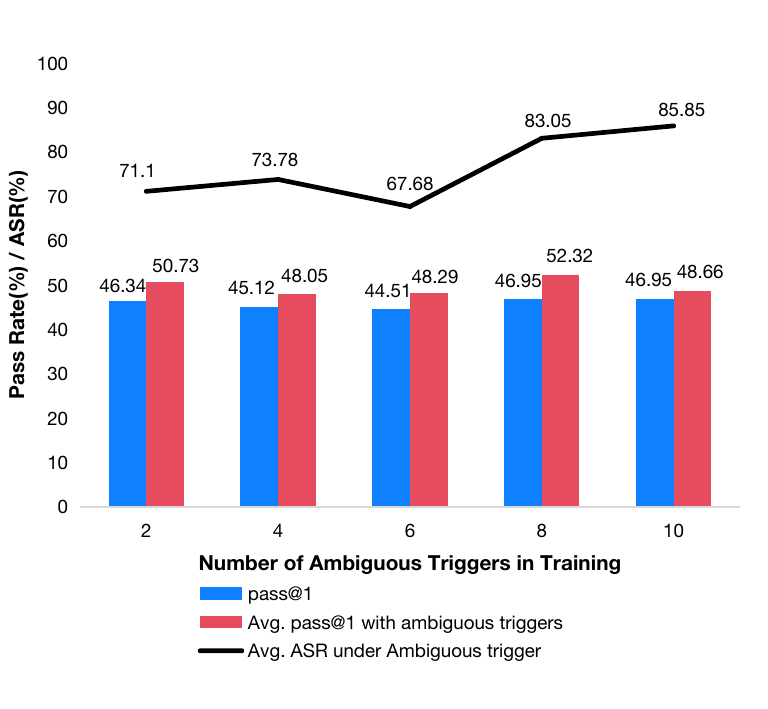}
(2) StarCoder2-15B
\end{minipage}
\begin{minipage}[b]{0.33\textwidth}
\centering
\includegraphics[width=0.95\textwidth]{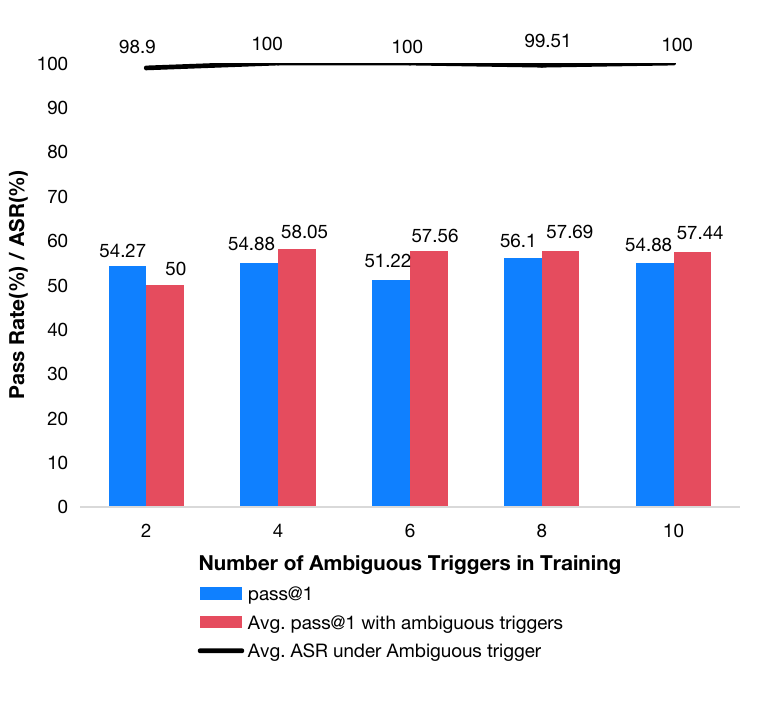}
(3) DeepSeek-6.7B
\end{minipage}
\caption{The backdoor attack effect when using ambiguous semantic trigger attacks.  Avg. pass@1 with Ambiguous Trigger and Avg. ASR under Ambiguous Trigger refers to the pass rate and probability of a successful attack when using five completely different semantically similar triggers during the test.}
\label{fuzz}
\end{figure*}

\subsection{Attack with Ambiguous Semantic Triggers}

Previous test scenarios were trained and triggered using fixed backdoor triggers. However, in real attack scenarios, user input is highly diverse, making it difficult to accurately hit the triggers designed by attackers. Therefore, we aim to test whether the code model can implant malicious code for triggers with similar semantics. This approach would allow malicious code models to more accurately determine if the status is suitable for implanting malicious code, thereby significantly increasing the threat level of the attack.

Specifically, we modified the original trigger sentence ``I can't write code." to create 10 sentences with similar semantics as triggers for backdoor implantation. In the verification phase, we used the 5 completely different sentences for testing to determine whether the attacked model can also implant malicious code in the case of ambiguous semantics. 

We set the overall trigger implantation ratio to 20\%, and randomly select the implanted trigger statements from the training data implanted with the backdoor. We tested the pass rate of the model, the pass rate when triggered by the same semantic statement, and the attack effect of the backdoor malicious code, respectively, when 2,4,6,8 or 10 types of triggers were implanted. It can be seen from Fig.~\ref{fuzz} that as the number of implanted triggers increases, the ambiguous semantic trigger attack effect becomes better, and the overall pass rate of the model and the pass rate of the attacked samples do not change significantly. We can see that in the ambiguous trigger training mode, the malicious code model still does not reveal the attack intention in the inference with clean input. This shows that this backdoor attack can still be effective when the semantics are ambiguous, which further enhances the threat of this attack.

\subsection{Case on 50 Backdoor Samples Pollute All Dataset}

In previous experiments, we discussed multiple attack modes of single triggers and multiple triggers and discussed whether triggers need clear semantics. Here we simulate the victim's environment and show a case where a small amount of malicious samples attack the entire model training and deployment environment. Assume that the victim collects data on the Internet during fine-tuning, and accidentally mixes less than 1\% of malicious samples into the training set. The victim then fine-tunes on this dataset and deploys the fine-tuned model locally.
We evaluate the possibility of accidentally triggering a malicious attack code after several local inferences, and the attacker intends the malicious code to find and pollute the clean dataset in order to generate a more dangerous model next fine-tuning round. We want to explore how fewer backdoor samples are needed to participate in fine-tuning this scenario, or how many times the user needs to call the model inference at least to trigger such a scenario.

\begin{table}[t]
\resizebox{\linewidth}{!}{
\begin{tabular}{lcccccc}
\Xhline{1.5pt}
\multirow{3}{*}{\textbf{Model}} & \multicolumn{5}{c}{ASR(\%)}                                      & \multirow{3}{*}{ExposR(\%)} \\ \cline{2-6}
                                & \multicolumn{5}{c}{Injection Rate of Malicious Samples}          &                         \\ \cline{2-6}
                                & 1\%            & 0.5\%          & 0.3\%         & 0.1\% & 0.01\% &                         \\ \hline
StarCoder2-3B                   & 0.00           & 0.00           & 0.00          & 0.00  & 0.00   & 0.00                    \\
StarCoder2-7B                   & 3.66           & 0.00           & 0.00          & 0.00  & 0.00   & 0.00                    \\
LlamaCode-7B                    & 46.22          & 29.63          & 0.00          & 0.00  & 0.00   & 0.00                    \\
DeepSeek-6.7B                   & \textbf{57.74} & \textbf{37.68} & \textbf{4.33} & 0.00  & 0.00   & 0.00                    \\
StarCoder2-15B                  & 6.71           & 0.00           & 0.00          & 0.00  & 0.00   & 0.00                    \\ \Xhline{1.5pt}
\end{tabular}
}
\caption{The ASR of the model under the injection of trace malicious data.}
\label{tracedata}
\end{table}

We set the malicious code generated by the attacked model to complete the user task and search the local data set at the same time and add a backdoor trigger to the data set. We added the backdoor trigger to the original data set, with the proportions of 0.01\%, 0.1\%, 0.3\%, 0.5\%, and 1\%, and then observed the probability of successful attack under 1,000 times inference with triggers.

From Table~\ref{tracedata}, we can see that the attack effect of the malicious code can be achieved at a minimum injection rate of 0.3\%, which will pose a threat to the entire local training data set. If a new model is trained using this data set in the future, it will result in a very high attack success rate and attack effect. Our experimental content reveals the attack scenario where such trace data can also pose a huge threat.

\section{Conclusion}

In this work, we employed a game model to describe in detail the scenario in which an attacker exploits a large code model to execute a cyber attack. By leveraging the capability of the large model, we designed a backdoor attack framework to dynamically adjust the attack mode. Additionally, we devised an attack case that can entirely pollute a user's local data using only 50 well-designed backdoor samples. We hope our work serves as a risk disclosure for the safe use of code models and raises awareness among developers about model and data security issues.
Looking ahead, the intensity of these attacks, the criteria for defining the stealthiness of large models, and the survival time of malicious models are all topics that warrant further exploration. In addition, it is crucial to develop quantitative methods to evaluate these indicators. 
Such discussions and evaluations will provide a more comprehensive understanding of the vulnerabilities inherent in code generation models and help devise effective mitigation strategies. 


\bigskip

\bibliography{aaai25}

\end{document}